# Anomaly Generation Using Generative Adversarial Networks in Host-Based Intrusion Detection

Milad Salem
Dept of Computer Engineering
University of Central Florida
Orlando, FL, USA
miladsalem@knights.ucf.edu

Shayan Taheri
Dept of Computer Engineering
University of Central Florida
Orlando, FL, USA
shayan.taheri@knights.ucf.edu

Jiann Shiun Yuan
Dept of Computer Engineering
University of Central Florida
Orlando, FL, USA
Jiann-Shiun.Yuan@ucf.edu

*Abstract*—Generative adversarial networks have been able to generate striking results in various domains. This generation capability can be general while the networks gain deep understanding regarding the data distribution. In many domains, this data distribution consists of anomalies and normal data, with the anomalies commonly occurring relatively less, creating datasets that are imbalanced. The capabilities that generative adversarial networks offer can be leveraged to examine these anomalies and help alleviate the challenge that imbalanced datasets propose via creating synthetic anomalies. This anomaly generation can be specifically beneficial in domains that have costly data creation processes as well as inherently imbalanced datasets. One of the domains that fits this description is the host-based intrusion detection domain. In this work, ADFA-LD dataset is chosen as the dataset of interest containing system calls of small foot-print next generation attacks. The data is first converted into images, and then a Cycle-GAN is used to create images of anomalous data from images of normal data. The generated data is combined with the original dataset and is used to train a model to detect anomalies. By doing so, it is shown that the classification results are improved, with the AUC rising from 0.55 to 0.71, and the anomaly detection rate rising from 17.07% to 80.49%. The results are also compared to SMOTE, showing the potential presented by generative adversarial networks in anomaly generation.

## I. Introduction

Since their first introduction in [1], generative adversarial networks (GANs) have been widely used in several diverse fields such as Image enhancement and drug discovery. The success of GANs in generating data originating from different distributions and types, shows the generality of the concept of GANs as well as their capability to learn nuances in the data distribution. Via using this potential, one can generate data instances that belong to the original data distribution, while avoiding the generation of data instances that come from the mean of the data distribution and lack realistic details.

With their learnt understanding of the data distribution, GANs' generative power can be used to solve the problem of imbalance in the distribution, specifically the imbalance between anomaly and normal data instances. While this imbalance in inherently present in many datasets, its existence is problematic. If a model is trained on a dataset with imbalanced anomalies to normal data ratio, it can favor the majority class of normal data, due to the fact that it simply is not presented with enough data regarding the anomaly distribution. Via training GANs to solve this problem, the task of understanding the difference between the normal and anomalous data distributions can be given to the GANs. This enables the GANs to generate realistic anomalies, adding to the data in the anomaly distribution, allowing an additional model to benefit from more available data.

This anomaly generation is beneficial if, firstly, the anomalies in a domain are rarely seen, and secondly, generating or recording anomalies in the real world in this domain is expensive. One domain that satisfies these requirements is the Host based intrusion detection. In this domain, the normal data is abundant compared to anomalous intrusion data, while creating these intrusions in a diverse manner is costly. This has caused the fact that few databases exist with recordings regarding host-based data of intrusions.

In this work, we use a Cycle-GAN to learn the transformations between normal and anomalous host-based data. Having done so, we learn the transformation of normal data to anomalous data and use it to generate anomalies from the normal data. The generated anomalies are then added to the anomaly distribution and are shown to help detect unseen anomalies after training an artificial neural network (ANN). Therefore, the main goal of this work is to improve the performance of host-based intrusion detection systems through generating anomalies and gaining better understanding of the anomaly distribution using GANs.

In the related literature, GANs have been used to generate anomalies in [2] and [3], yet this work, to the authors' knowledge, is the first work to leverage the existence of a huge number of normal instances and transforming them into anomalies.

## II. Background

### A. Host-Based Intrusion Detection Systems

Host-based intrusion detection system (HIDS) is a defensive tool against intrusions after the cyber-attacks have passed through the network security defense measures, such as firewall or network intrusion detection systems, and are on a given host. The main purpose of having this system is internal monitoring through screening the information collected from a computer, node, or device to determine whether the host has been compromised. The intrusion can be detected by finding abnormal behavior in host-based information such as system calls, system logs, applications actions, and host traffic [4]. Specifically, the system calls are considered the most expensive source of audit data in Unix-based systems. They are the basic means for identification of computing processes at the higher level of computer hierarchy that can be executed by any process. Monitoring these processes helps find abnormalities.

An HIDS usually consists of three sections; 1) Data source containing the recorded host-based audit data, 2) feature retrieval, a section responsible for abstraction of the data and extracting meaningful features from it, and 3) decision engine that determines if an instance is anomalous or not [5]. In this work, we focus on improving the performance of the decision engine by appending generated anomalies to the data source.

*B. Dataset in Use*

ADFA-LD first introduced in [6], is a host-based intrusion data source containing recorded system calls of compromised and non-compromised hosts. This dataset contains data regarding six new generation attacks such as user to root, brute force, exploits, and backdoors [7], which combine their anomalous behavior with normal behavior. The ADFA-LD dataset contains 5951 total instances, with 5205 normal instances and 746 anomalous instances. This creates an imbalanced dataset with a ratio of 1:7 for normal to anomalous data. As mentioned in the previous section, a feature retrieval function is needed to extract the features. In this work, the abstracted data from the approach integer data zero-watermark assisted system calls abstraction is acquired from [5]. Therefore, the feature extraction is not performed and the dataset in use is already in integer format.

*C. Cycle-GAN*

The modifications and alterations implemented on the structure of GANs have resulted in different types of GANs. While the generator in the original GAN has random noise as input, in Cycle-GAN created in [8], the input is an image from a given distribution, while output is an image belonging to another distribution. The structure of Cycle-GAN is shown in Fig. 1.

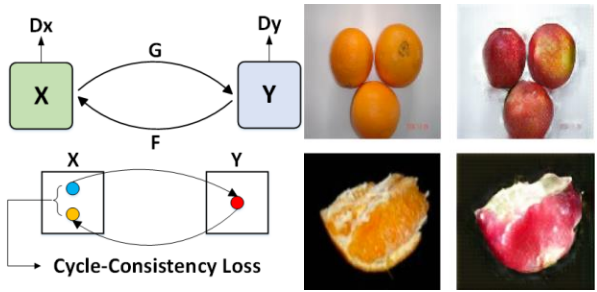

Fig. 1. The Cycle-GAN structure (left) and samples of the transformation (right [9]).

In this structure, the Cycle-GAN learns the transformation between instances from distribution X to Y (function G) and vice versa (function F). While $D_x$ and $D_y$ are responsible for the realism of the outcome of these transformations and reassure that the generated instance belongs to their respective distribution. The structure of the Cycle-GAN consists of two convolutions, several residual blocks, and two convolutions for the generator and $70 \times 70$ PatchGANs for the discriminator [8].

Cycle-GAN operates by transforming instances from a distribution to another and then transforming them back to the original distribution. With this approach if cycle-consistency is implemented, functions G and F learn to transform data from one distribution to another without losing considerable information in the transformation. Therefore, the overall structure of the data instance is saved, while changes occur in the values of the image. This can be seen in Fig. 1, where oranges are transformed into apples, with the overall structure being preserved, while defects in the orange are mapped to defects in the apple, and the skin and inside of the orange are mapped to their respective parts in the generated apple.

Cycle-GAN is compatible with generation from ADFA-LD dataset in two manners: 1) the structure of the data is saved when transforming between the two data distributions that are governed by the same structure and feature abstraction, 2) the integer-based data is compatible with the nature of Cycle-GAN which works with images made of integers from 0 to 255.

III. METHODOLOGY

*A. Anomaly Generation and Classification System*

The one-dimensional host-based intrusion data from the ADFA-LD dataset is first partitioned into training and test sets and transformed into two-dimensional images. To generate anomalies, we aim to transform normal data into anomalous data, therefore, a part of the normal instances in the training set is extracted and is named "Template Data", which we use as references that undergo the normal to anomaly transformations. The anomaly generation and evaluation system is shown in Fig. 2.

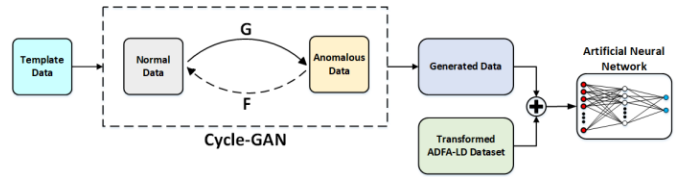

Fig. 2. Anomaly generation and classification system

Firstly, the Cycle-GAN is trained using normal data as one distribution and anomalous data as the other. Having done so, Cycle-GAN learns to transform normal data to anomalous data. Then the template data is fed to the function G, resulting in the generation of synthetic anomalous data. This generated data is appended to the imbalanced training data, creating a balance between the normal and anomalous classes of the dataset. Subsequently, the balanced dataset is used to train a Multilayer Perceptron (MLP), and the classification is evaluated using the test set.

## B. Data Transformation and Partitioning

In this work, the host-based intrusion data is transformed into images to leverage the capabilities of Cycle-GAN. The fact that the dataset of interest is integer-based facilitates the data transformation into images, by lowering the loss of quantization and normalization. The Cycle-GAN takes images of fixed sizes as input, thus an optimal size for the image should be defined that can contain most of the data, while preserve their fine-grained changes. The lengths of the data instances are various ranging from 75 to 4494, however, 89% of the data are shorter that 1024 integers. Therefore, all data is shorter than 1024 is extended to this length by adding the value of 255 and the data that are longer than 1024 are discarded. The transformation occurs by putting the 1024 integers into a 32x32 image in a row-wise manner. The results of the transformation are seen in Fig. 3.

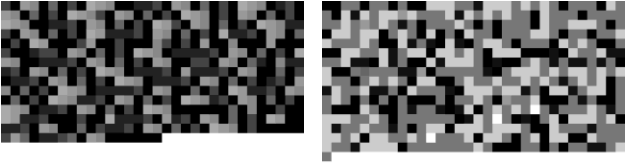

Fig. 3. Transformed data into images, normal (right) and anomalous (left).

The white sections visible in Fig. 3 denote the part of the data that was extended and replaced by 255 in order to fill the image. The patterns created in these images for normal and anomalous classes will enable the discrimination between the two classes.

To begin the classification process, the dataset needs to be partitioned into training and test. 30% of the data is chosen as the test set and the remaining as the imbalanced training set. To avoid leakage of data to the generation and training process and to have fair classification, the test set is isolated from the rest of the process and is never involved until the evaluation process. The test data contains 446 normal instances and 82 anomalous instances.

The Cycle-GAN is trained using the imbalanced training set, learning to transform normal images to anomalous images and vice versa. However, the MLP is trained using a balanced training set, by combining the generated images with the imbalanced training set. The training data include 4150 instances of normal data and 598 instances of anomalous data. In order to balance the dataset, 3552 anomaly data need to be generated. Therefore, 3552 normal data are extracted from the dataset randomly, and create the template data, which are transformed via Cycle-GAN.

## C. Learning Rate

Since Cycle-GAN has parameters and structure optimized for handling images, the structure and hyper-parameters are not changed in this work. However, since the complexity and the data distribution of the images in this work differ from that of the Cycle-GAN, the learning rate is changed to aid the classification process. The change of the learning rate is shown in Fig. 4.

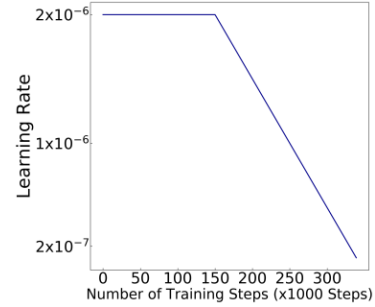

Fig. 4. Learning rate through training steps.

As seen in Fig. 4, the learning rate remains $2x10^{-6}$ for 150,000 steps, after which it linearly decreases to $2x10^{-7}$ in the next 200,000 steps. This range is determined trial by and error, delivering results that are relatively fast, yet have good quality.

## D. Evaluation Method

Having balanced the training dataset, an MLP is trained and evaluated. This MLP consists of two layers with the size of 100 and 20, respectively. To evaluate the results of the classification the area under the ROC curve (AUC) is used, which is a prevalent metric used in similar works that focus on classifying imbalanced datasets.

Furthermore, two additional MLPs are trained using imbalanced data and balanced data via Synthetic Minority Over-Sampling Technique (SMOTE). SMOTE serves as the most prevalent approach for minority generation, creating new synthetic data via iteratively over-sampling the minority group. In this work, SMOTE generates new anomalous data by over-sampling the anomalies in the dataset, while our approach creates new data by understanding the difference and the transition between anomaly and normal data distributions.

## IV. RESULTS

### A. Results of Training the Cycle-GAN

The implementation of the Cycle-GAN with TensorFlow backend was acquired from [9]. After changing the learning rate, the model was trained for 330,000 steps. The values of the loss function for F, G, $D_x$, and $D_y$ were recorded every 10,000 steps and are shown in Fig. 5.

As it is visible in Fig. 5, the G function shows more volatility in training than the F function. The G function transforms the input to the anomaly distribution, which is a less populated distribution. Therefore, compared to the F function, this function is presented with less information and acts in a more volatile manner. The same case can be seen when comparing $D_x$ and $D_y$, with $D_x$ being more volatile when judging the realism of the generated images based from the fewer anomaly images.

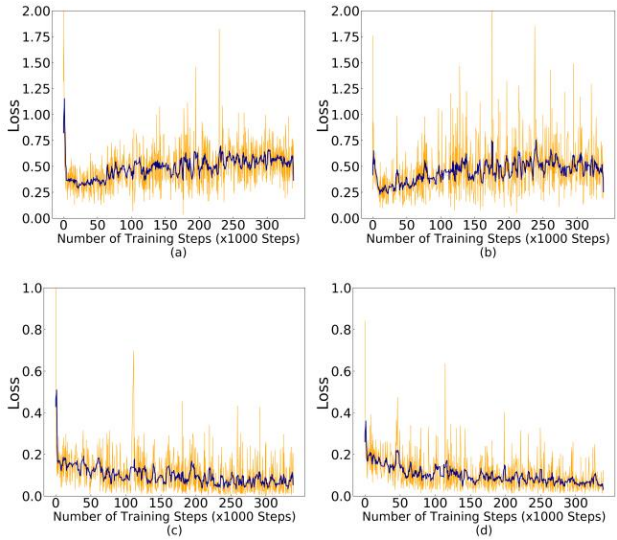

Fig. 5. Values of the loss functions through training steps for the functions of; a) F, b) G, c) $D_x$, d) $D_y$.

## B. The Evolution of Generated Images

The volatility in the loss functions seen in Fig. 1 proposes a challenge as to when to stop the training of the GAN. To overcome this challenge the model of the Cycle-GAN was saved every 10000 steps and is evaluated later. The saved models are exported into graphs and the template data, which consists of normal data, is fed to these graphs in order to generate synthetic anomalous data. The transformation that occurs on a normal image via each model is shown in Fig. 6.

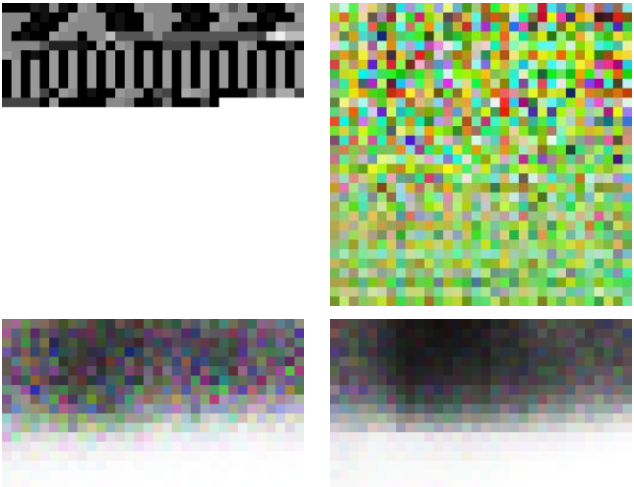

Fig. 6. The evolution of generated images in different steps; original image (upper left), 0 steps (upper right), 10,000 steps (lower left), 150,000 steps (lower right).

As seen in Fig. 6, the model starts with no knowledge of the data distributions and generates the value of each pixel randomly. At 10,000 steps, the model has learnt the concept of preserving the white section of the image and is generating images that are close to gray scale, similar to the original data. At 150,000 steps the generated image contains a sharper border above the white section. Cycle-GAN layers contain 2-D convolutional filters, which use the surrounding area of the border line to reconstruct the original border line and maintain cycle-consistency. Therefore, information is stored around the border line to help the reconstruction and this border line is never perfect. After 60,000 steps, the generated images preserve their overall patterns with various changes in the pixel values.

## C. Evaluation Results

After the data is generated with each individual graph, the MLP is trained using the original data combined with the generated data. The AUC is recorded for each model and is shown in Fig. 7.

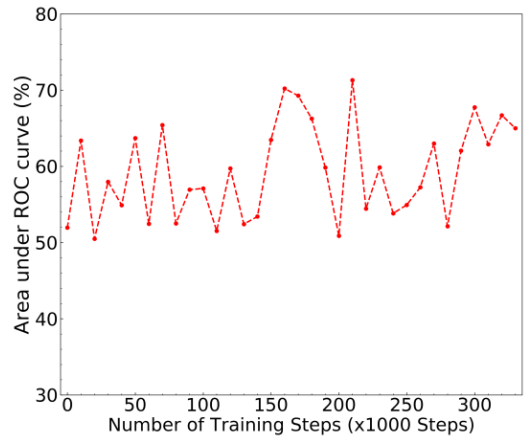

Fig. 7. AUC of different generation models created at different steps.

A few models deliver relatively high AUC and a few models perform similar to the model at step 0, which generates random results, therefore, the volatility of the Cycle-GAN is still visible in the final results in Fig. 7. The model that created the best result is the one saved at 210,000 steps.

The MLP is trained two more times, once using no generated data and once using SMOTE to balance the dataset instead of Cycle-GAN. The results are reflected in Table I.

TABLE I. COMPARISON OF CLASSIFICATION RESULTS

| Approach | TP | TN | FP | FN | Recall | F1 | AUC |
|---|---|---|---|---|---|---|---|
| Imbalanced Data | 14 | 415 | 31 | 68 | 17.07 | 22.05 | 55.06 |
| SMOTE | 67 | 250 | 196 | 15 | 81.71 | 38.84 | 68.89 |
| Cycle-GAN | 66 | 277 | 169 | 16 | 80.49 | 41.64 | 71.30 |

As it is visible form Table I, classification on imbalanced data results in a model that mostly predicts the input as normal, since normal is the majority class and the model is not given enough information to discriminate between the two classes.

When comparing the approaches that have balanced datasets, our approach outperforms SMOTE with lower false positives and higher AUC.

Our approach has room for improvement via using 1-D convolutional neural networks and customizing the structure of the neural network to be compatible with the dataset of interest. Despite the lack of these customizations, Cycle-GAN was able to deliver better results than SMOTE, which is a predominant method for balancing the data. This superiority originates from the fact that Cycle-GAN can learn different concepts in the data distributions and transforms the data in a way that it remains conceptually correct. These results show the capability of Cycle-GAN in anomaly generation.

## V. Conclusion

This work, to the authors' knowledge, is the first attempt to use Cycle-GAN to generate anomalous data via transforming the normal data. The abstracted ADFA-LD dataset, which contains data from host-based intrusions, is first transformed into images. Part of the images from the normal class are fed to the Cycle-GAN as template data and are transformed into anomalies. These synthetic anomalies were then combined with the original dataset and are fed to an MLP. The results show an improvement in the effectiveness of the classification results, increasing the AUC from 0.55 to 0.71, and increasing the percentage of detected unseen anomalies from 17.07% to 80.49%. This approach was able to outperform SMOTE, despite being in its early stages of customization, showing the potential of generative adversarial networks in the anomaly generation domain.